\documentclass[10pt]{article}
\usepackage{amsmath}
\usepackage{amssymb}
\usepackage{amsfonts}
\usepackage{algorithm}
\usepackage{algorithmic}
\usepackage[font={footnotesize,sf},labelfont={footnotesize,sf,bf},skip=2.5pt]{caption}
\usepackage{graphicx}
\usepackage{color}

\newcommand{\vect}[1]{\mathbf{#1}}

\newcommand{\rnn}{\mathcal{R}}
\newcommand{\inseq}{\vect{x}}
\newcommand{\stateseq}{\vect{s}}
\newcommand{\outseq}{\vect{y}}

\newcommand{\statef}{\mathcal{S}}

\newcommand{\loss}{\mathcal{L}}
\newcommand{\tloss}{\mathcal{P}}

\newcounter{mnote}

\newcommand{\ie}{i.e.\ }

\newcommand{\flabel}[1]{\label{fig:#1}}
\newcommand{\seclabel}[1]{\label{sec:#1}}
\newcommand{\tlabel}[1]{\label{tab:#1}}
\newcommand{\elabel}[1]{\label{eq:#1}}

\newcommand{\fref}[1]{\Cref{fig:#1}}
\newcommand{\sref}[1]{\Cref{sec:#1}}
\newcommand{\tref}[1]{\Cref{tab:#1}}
\newcommand{\eref}[1]{\Cref{eq:#1}}

\newcommand*\idx[2][]
{ 
\def\next{#1}%
\ifx\empty\next
  (#2)
\else
  (#1, #2)
\fi
}
\newcommand*\elt[3][]
{ 
\def\next{#1}%
\ifx\empty\next
  #2\idx{#3}
\else
  #1\idx{#2,#3}
\fi
}
\newcommand*\pd[3][]
{ 
\def\next{#1}%
\ifx\empty\next
  \frac{\partial#2}{\partial #3}
\else
  \frac{{\partial^{#1} #2}}{\partial#3^{#1}}
\fi
}

\newcommand{\figdir}{fig/}
\newcommand{\capt}[2]{\caption[#1.]{\textbf{#1.}#2}}

\newcommand{\figt}[5]
{
\begin{figure}
\begin{center}
\includegraphics[width=#3\columnwidth]{\figdir/#1}
\end{center}
\capt{#4}{#5}
\flabel{#2}
\end{figure}
}

\newcommand{\plotwidth}{1}
\newcommand{\barwidth}{1}

\usepackage{hyperref}
\usepackage{cleveref}

\author{Alex Graves\\
Google DeepMind\\
\texttt{gravesa@google.com}
}
\date{}
\title{
\rule[0.4cm]{\textwidth}{2pt}
{\bf Adaptive Computation Time\\for Recurrent Neural Networks}
\rule{\textwidth}{2pt} 
}

\begin{document}
\maketitle

\begin{center}
{\bf Abstract} 
\end{center}
This paper introduces \emph{Adaptive Computation Time} (ACT), an algorithm that allows recurrent neural networks to learn how many computational steps to take between receiving an input and emitting an output.
ACT requires minimal changes to the network architecture, is deterministic and differentiable, and does not add any noise to the parameter gradients.
Experimental results are provided for four synthetic problems: determining the parity of binary vectors, applying binary logic operations, adding integers, and sorting real numbers.
Overall, performance is dramatically improved by the use of ACT, which successfully adapts the number of computational steps to the requirements of the problem.
We also present character-level language modelling results on the Hutter prize Wikipedia dataset.
In this case ACT does not yield large gains in performance; however it does provide intriguing insight into the structure of the data, with more computation allocated to harder-to-predict transitions, such as spaces between words and ends of sentences.
This suggests that ACT or other adaptive computation methods could provide a generic method for inferring segment boundaries in sequence data.

\section{Introduction}
The amount of time required to pose a problem and the amount of thought required to solve it are notoriously unrelated.
Pierre de Fermat was able to write in a margin the conjecture (if not the proof) of a theorem that took three and a half centuries and reams of mathematics to solve~\cite{wiles95fermat}.
More mundanely, we expect the effort required to find a satisfactory route between two cities, or the number of queries needed to check a particular fact, to vary greatly, and unpredictably, from case to case.
Most machine learning algorithms, however, are unable to dynamically adapt the amount of computation they employ to the complexity of the task they perform.

For artificial neural networks, where the neurons are typically arranged in densely connected layers, an obvious measure of computation time is the number of layer-to-layer transformations the network performs.
In feedforward networks this is controlled by the network \emph{depth}, or number of layers stacked on top of each other.
For recurrent networks, the number of transformations also depends on the length of the input sequence --- which can be padded or otherwise extended to allow for extra computation.
The evidence that increased depth leads to more performant networks is by now inarguable~\cite{dahl2012speech,ciresan2012multicolumn,krizhevsky2012imagenet,graves2013speech}, and recent results show that increased sequence length can be similarly beneficial~\cite{sukhbaatar2015end,vinyals2015order,reed2015npi}.
However it remains necessary for the experimenter to decide \emph{a priori} on the amount of computation allocated to a particular input vector or sequence.
One solution is to simply make every network very deep and design its architecture in such a way as to mitigate the \emph{vanishing gradient problem}~\cite{hochreiter2001gradient} associated with long chains of iteration~\cite{srivastava2015training,kalchbrenner2015grid}.
However in the interests of both computational efficiency and ease of learning it seems preferable to dynamically vary the number of steps for which the network `ponders' each input before emitting an output.
In this case the effective depth of the network at each step along the sequence becomes a dynamic function of the inputs received so far.

The approach pursued here is to augment the network output with a sigmoidal \emph{halting unit} whose activation determines the probability that computation should continue.
The resulting \emph{halting distribution} is used to define a mean-field vector for both the network output and the internal network state propagated along the sequence.
A stochastic alternative would be to halt or continue according to binary samples drawn from the halting distribution---a technique that has recently been applied to scene understanding with recurrent networks~\cite{eslami2016attend}. 
However the mean-field approach has the advantage of using a smooth function of the outputs and states, with no need for stochastic gradient estimates.
We expect this to be particularly beneficial when long sequences of halting decisions must be made, since each decision is likely to affect all subsequent ones, and sampling noise will rapidly accumulate (as observed for policy gradient methods~\cite{williams1995gradient}). 

A related architecture known as \emph{Self-Delimiting Neural Networks}~\cite{schmidhuber2012self,srivastava2013first} employs a halting neuron to end a particular update within a large, partially activated network; in this case however a simple activation threshold is used to make the decision, and no gradient with respect to halting time is propagated.
More broadly, learning when to halt can be seen as a form of \emph{conditional computing}, where parts of the network are selectively enabled and disabled according to a learned policy~\cite{bengio2015conditional,denoyer2014deep}.

We would like the network to be parsimonious in its use of computation, ideally limiting itself to the minimum number of steps necessary to solve the problem.
Finding this limit in its most general form would be equivalent to determining the Kolmogorov complexity of the data (and hence solving the halting problem)~\cite{li2013introduction}.
We therefore take the more pragmatic approach of adding a time cost to the loss function to encourage faster solutions.
The network then has to learn to trade off accuracy against speed, just as a person must when making decisions under time pressure.
One weakness is that the numerical weight assigned to the time cost has to be hand-chosen, and the behaviour of the network is quite sensitive to its value.

The rest of the paper is structured as follows: the Adaptive Computation Time algorithm is presented in \sref{act}, experimental results on four synthetic problems and one real-world dataset are reported in \sref{results}, and concluding remarks are given in \sref{conclusion}.

\section{Adaptive Computation Time}\seclabel{act}
Consider a recurrent neural network $\rnn$ composed of a matrix of \emph{input weights} $W_x$, a parametric \emph{state transition model} $\statef$, a set of \emph{output weights} $W_y$ and an \emph{output bias} $b_y$. 
When applied to an input sequence $\inseq = (x_1,\ldots,x_T)$, $\rnn$ computes the \emph{state sequence} $\stateseq = (s_1,\ldots,s_T)$ and the output sequence $\outseq = (y_1,\ldots,y_T)$ by iterating the following equations from $t=1$ to $T$:
\begin{align}
s_t &= \statef(s_{t-1}, W_x x_t)\\
y_t &= W_y s_t + b_y
\end{align} 
The state is a fixed-size vector of real numbers containing the complete dynamic information of the network.
For a standard recurrent network this is simply the vector of hidden unit activations.
For a Long Short-Term Memory network (LSTM)~\cite{hochreiter1997lstm}, the state also contains the activations of the memory cells.
For a memory augmented network such as a Neural Turing Machine (NTM)~\cite{graves2014ntm}, the state contains both the complete state of the controller network and the complete state of the memory.
In general some portions of the state (for example the NTM memory contents) will not be visible to the output units; in this case we consider the corresponding columns of $W_y$ to be fixed to 0.

\emph{Adaptive Computation Time} (ACT) modifies the conventional setup by allowing $\rnn$ to perform a variable number of state transitions and compute a variable number of outputs at each input step.
Let $N(t)$ be the total number of updates performed at step $t$.
Then define the \emph{intermediate state sequence} $(s^1_t,\ldots,s^{N(t)}_t)$ and \emph{intermediate output sequence} $(y^1_t,\ldots,y^{N(t)}_t)$ at step $t$ as follows
\begin{align}
\elabel{rnn_hidden}
s^n_t &= \begin{cases}\statef(s_{t-1}, x^1_t)\text{ if }n=1\\\statef(s^{n-1}_{t}, x^n_t)\text{ otherwise}\end{cases}\\
y^n_t &= W_y s^n_t + b_y
\end{align}
where $x^n_t = x_t + \delta_{n,1}$ is the input at time $t$ augmented with a binary flag that indicates whether the input step has just been incremented, allowing the network to distinguish between repeated inputs and repeated computations for the same input.
Note that the same state function is used for all state transitions (intermediate or otherwise), and similarly the output weights and bias are shared for all outputs.
It would also be possible to use different state and output parameters for each intermediate step; however doing so would cloud the distinction between increasing the number of parameters and increasing the number of computational steps.
We leave this for future work.

To determine how many updates $\rnn$ performs at each input step an extra sigmoidal \emph{halting} unit $h$ is added to the network output, with associated weight matrix $W_h$ and bias $b_h$:
\begin{equation}\elabel{halting_act}
h^n_t = \sigma\left(W_h s^n_t + b_h\right)
\end{equation}
As with the output weights, some columns of $W_h$ may be fixed to zero to give selective access to the network state.
The activation of the halting unit is then used to determine the \emph{halting probability} $p^n_t$ of the intermediate steps:
\begin{align}\elabel{halting_prob}
p^n_t = \begin{cases}R(t)\text{ if }n=N(t)\\h^n_t\text{ otherwise}\end{cases}
\end{align}
where
\begin{equation}\elabel{epsilon}
N(t) =\min\{n':\sum_{n=1}^{n'} h^n_t >= 1-\epsilon\}
\end{equation}
the \emph{remainder} $R(t)$ is defined as follows
\begin{equation}\elabel{remainder}
R(t) = 1 - \sum_{n=1}^{N(t)-1} h^n_t
\end{equation}
and $\epsilon$ is a small constant (0.01 for the experiments in this paper), whose purpose is to allow computation to halt after a single update if $h^1_t >= 1-\epsilon$, as otherwise a minimum of two updates would be required for every input step.
It follows directly from the definition that $\sum_{n=1}^{N(t)}p^n_t = 1$ and $0 \leq p^n_t\leq 1\ \forall n$, so this is a valid probability distribution.
A similar distribution was recently used to define differentiable \emph{push} and \emph{pop} operations for neural stacks and queues~\cite{grefenstette2015learning}.

At this point we could proceed stochastically by sampling $\hat{n}$ from $p^n_t$ and setting $s_t = s^{\hat{n}}_t$, $y^t = y^{\hat{n}}_t$.
However we will eschew sampling techniques and the associated problems of noisy gradients, instead using $p^n_t$ to determine mean-field updates for the states and outputs:
\begin{equation}
\elabel{mean_field}
s_t = \sum_{n=1}^{N(t)} p^n_t s^n_t \qquad
y_t = \sum_{n=1}^{N(t)} p^n_t y^n_t
\end{equation}
The implicit assumption is that the states and outputs are approximately linear, in the sense that a linear interpolation between a pair of state or output vectors will also interpolate between the properties the vectors embody.
There are several reasons to believe that such an assumption is reasonable.
Firstly, it has been observed that the high-dimensional representations present in neural networks naturally tend to behave in a linear way~\cite{sutskever2014sequence,le2014distributed}, even remaining consistent under arithmetic operations such as addition and subtraction~\cite{mikolov2013distributed}.
Secondly, neural networks have been successfully trained under a wide range of adversarial regularisation constraints, including sparse internal states~\cite{olshausen1996emergence}, stochastically masked units~\cite{srivastava2014dropout} and randomly perturbed weights~\cite{an1996effects}.
This leads us to believe that the relatively benign constraint of approximately linear representations will not be too damaging.
Thirdly, as training converges, the tendency for both mean-field and stochastic latent variables is to concentrate all the probability mass on a single value.
In this case that yields a standard RNN with each input duplicated a variable, but deterministic, number of times, rendering the linearity assumption irrelevant.

A diagram of the unrolled computation graph of a standard RNN is illustrated in \fref{rnn_graph}, while \fref{act_graph} provides the equivalent diagram for an RNN trained with ACT.

\figt{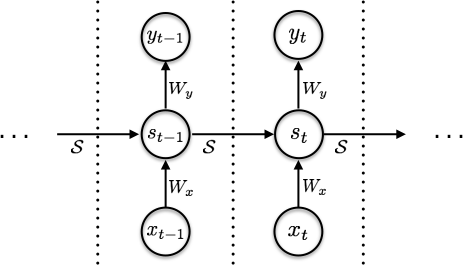}{rnn_graph}{0.45}{RNN Computation Graph}{ An RNN unrolled over two input steps (separated by vertical dotted lines). The input and output weights $W_x,W_y$, and the state transition operator $\mathcal{S}$ are shared over all steps.}

\figt{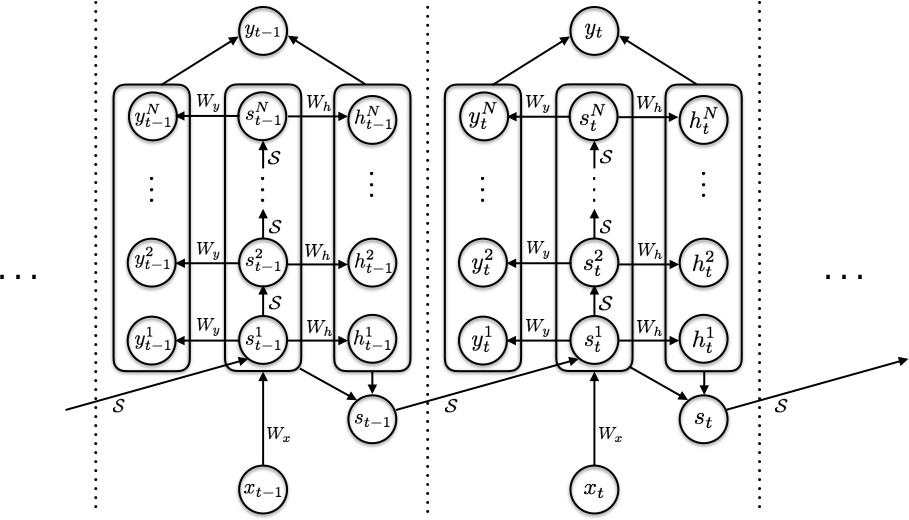}{act_graph}{0.8}{RNN Computation Graph with Adaptive Computation Time}{ The graph is equivalent to \fref{rnn_graph}, only with each state and output computation expanded to a variable number of intermediate updates. Arrows touching boxes denote operations applied to all units in the box, while arrows leaving boxes denote summations over all units in the box.}

\subsection{Limiting Computation Time}
If no constraints are placed on the number of updates $\rnn$ can take at each step it will naturally tend to `ponder' each input for as long as possible (so as to avoid making predictions and incurring errors).
We therefore require a way of limiting the amount of computation the network performs.
Given a length $T$ input sequence $\inseq$, define the \emph{ponder sequence} $(\rho_1,\dots,\rho_T$) of $\rnn$ as
\begin{equation}
\rho_t = N(t) + R(t)
\end{equation}
and the \emph{ponder cost} $\tloss(\inseq)$ as
\begin{equation}
\tloss(\inseq) = \sum_{t=1}^T \rho_t
\end{equation} 
Since $R(t) \in (0,1)$, $\tloss(\inseq) $ is an upper bound on the (non-differentiable) property we ultimately want to reduce, namely the total computation $\sum_{t=1}^T N(t)$ during the sequence\footnote{For a stochastic ACT network, a more natural halting distribution than the one described in \eref{halting_prob,eq:epsilon,eq:remainder} would be to simply treat $h^n_t$ as the probability of halting at step $n$, in which case $p^n_t = h^n_t\prod_{n'=1}^{n-1}(1-h^{n'}_t)$. One could then set $\rho_t = \sum_{n=1}^{N(t)} n p^n_t$ --- \ie the expected ponder time under the stochastic distribution. However experiments show that networks trained to minimise \emph{expected} rather than \emph{total} halting time learn to `cheat' in the following ingenious way: they set $h^1_t$ to a value just below the halting threshold, then keep $h^n_t = 0$ until some $N(t)$ when they set $h^{N(t)}_t$ high enough to ensure they halt. In this case $p^{N(t)}_t \ll p^1_t$, so the states and outputs at $n=N(t)$ have much lower weight in the mean field updates (\eref{mean_field}) than those at $n=1$; however by making the magnitudes of the states and output vectors much larger at $N(t)$ than $n=1$ the network can still ensure that the update is dominated by the final vectors, despite having paid a low ponder penalty.}.

We can encourage the network to minimise $\tloss(\inseq)$ by modifying the sequence loss function $\loss(\inseq,\outseq)$ used for training:
\begin{equation}\elabel{act_loss}
\hat{\loss}(\inseq,\outseq) =\loss{(\inseq,\outseq)} + \tau \tloss(\inseq)
\end{equation}
where $\tau$ is a \emph{time penalty} parameter that weights the relative cost of computation versus error.
As we will see in the experiments section the behaviour of the network is quite sensitive to the value of $\tau$, and it is not obvious how to choose a good value.
If computation time and prediction error can be meaningfully equated (for example if the relative financial cost of both were known) a more principled technique for selecting $\tau$ should be possible.

To prevent very long sequences at the beginning of training (while the network is learning how to use the halting unit) the bias term $b_h$ can be initialised to a positive value.
In addition, a hard limit $M$ on the maximum allowed value of $N(t)$ can be imposed to avoid excessive space and time costs.
In this case \eref{epsilon} is modified to
\begin{equation}\elabel{max}
N(t) = \min\{M, \min\{n':\sum_{n=1}^{n'} h^n_t >= 1-\epsilon\}\}
\end{equation}
\subsection{Error Gradients}
The ponder costs $\rho_t$ are discontinuous with respect to the halting probabilities at the points where $N(t)$ increments or decrements (that is, when the summed probability mass up to some $n$ either decreases below or increases above $1-\epsilon$).
However they are continuous away from those points, as $N(t)$ remains constant and $R(t)$ is a linear function of the probabilities.
In practice we simply ignore the discontinuities by treating $N(t)$ as constant and minimising $R(t)$ everywhere.

Given this approximation, the gradient of the ponder cost with respect to the halting activations is straightforward:
\begin{align}
\pd{\tloss(\inseq)}{h^n_t} &= \begin{cases} 0 \text{ if } n = N(t)\\-1 \text{ otherwise}\end{cases}
\end{align}
and hence
\begin{equation}
\elabel{halt_prob_ponder_derivs}
\pd{\hat{\loss}(\inseq,\outseq)}{h^n_t} = \pd{\loss(\inseq,\outseq)}{h^n_t} - \begin{cases} 0 \text{ if } n = N(t)\\\tau \text{ otherwise}\end{cases} 
\end{equation}
The halting activations only influence $\loss$ via their effect on the halting probabilities, therefore
\begin{equation}
\pd{\loss(\inseq,\outseq)}{h^n_t} = \sum_{n'=1}^{N(t)} \pd{\loss(\inseq,\outseq)}{p^{n'}_t}\pd{p^{n'}_t}{h^{n}_t}
\end{equation}
Furthermore, since the halting probabilities only influence $\loss$ via their effect on the states and outputs, it follows from \eref{mean_field} that
\begin{align}
\elabel{halt_prob_loss_derivs}
\pd{\loss(\inseq,\outseq)}{p^n_t} &= \pd{\loss(\inseq,\outseq)}{y_t}y^n_t + \pd{\loss(\inseq,\outseq)}{s_t}s^n_t
\end{align}
while, from \eref{halting_prob,eq:remainder}
\begin{align}
\elabel{halt_act_halt_prob_derivs}
\pd{p^{n'}_t}{h^{n}_t} &= \begin{cases} \delta_{n,n'} \text{ if } n' < N(t) \text{ and } n < N(t)\\-1 \text{ if } n' = N(t) \text{ and } n < N(t)\\0\text{ if } n = N(t)\end{cases}
\end{align}∫
Combining \eref{halt_prob_ponder_derivs,eq:halt_prob_loss_derivs,eq:halt_act_halt_prob_derivs} gives, for $n < N(t)$
\begin{align}
\pd{\hat{\loss}(\inseq,\outseq)}{h^n_t} 
&= \pd{\loss(\inseq,\outseq)}{y_t}\left(y^n_t - y^{N(t)}_t \right) + \pd{\loss(\inseq,\outseq)}{s_t}\left(s^n_t - s^{N(t)}_t \right) - \tau
\end{align}
while for $n=N(t)$
\begin{equation}
\pd{\hat{\loss}(\inseq,\outseq)}{h^{N(t)}_t} = 0
\end{equation} 
Thereafter the network can be differentiated as usual (e.g.\ with backpropagation through time~\cite{williams1995gradient}) and trained with gradient descent.
\section{Experiments}\seclabel{results}
We tested recurrent neural networks (RNNs) with and without ACT on four synthetic tasks and one real-world language processing task.
LSTM was used as the network architecture for all experiments except one, where a simple RNN was used.
However we stress that ACT is equally applicable to any recurrent architecture.

All the tasks were supervised learning problems with discrete targets and cross-entropy loss.
The data for the synthetic tasks was generated online and cross-validation was therefore not needed.
Similarly, the character prediction dataset was sufficiently large that the network did not overfit.
The performance metric for the synthetic tasks was the \emph{sequence error rate}: the fraction of examples where \emph{any} mistakes were made in the complete output sequence.
This metric is useful as it is trivial to evaluate without decoding. 
For character prediction the metric was the average log-loss of the output predictions, in units of bits per character.

Most of the training parameters were fixed for all experiments: Adam~\cite{kingma2014adam} was used for optimisation with a learning rate of $10^{-4}$, the Hogwild! algorithm~\cite{recht2011hogwild} was used for asynchronous training with 16 threads; the initial halting unit bias $b_h$ mentioned in \eref{halting_act} was 1; the $\epsilon$ term from \eref{epsilon} was 0.01.
The synthetic tasks were all trained for 1M iterations, where an iteration is defined as a weight update on a single thread (hence the total number of weight updates is approximately 16 times the number of iterations).
The character prediction task was trained for 10K iterations. 
Early stopping was not used for any of the experiments.

A logarithmic grid search over time penalties was performed for each experiment, with 20 randomly initialised networks trained for each value of $\tau$.
For the synthetic problems the range of the grid search was from $i \times 10^{-j}$ with integer $i$ in the range 1--10 and the exponent $j$ in the range 1--4.
For the language modelling task, which took many days to complete, the range of $j$ was limited to 1--3 to reduce training time (lower values of $\tau$, which naturally induce more pondering, tend to give greater data efficiency but slower wall clock training time).

Unless otherwise stated the maximum computation time $M$ (\eref{max}) was set to 100.
In all experiments the networks converged on learned values of $N(t)$ that were far less than $M$, which functions mainly as safeguard against excessively long ponder times early in training.

\subsection{Parity}
Determining the parity of a sequence of binary numbers is a trivial task for a recurrent neural network~\cite{schmidhuber96guessing}, which simply needs to implement an internal switch that changes sign every time a one is received.
For shallow feedforward networks receiving the entire sequence in one vector, however, the number of distinct input patterns, and hence difficulty of the task, grows exponentially with the number of bits.
We gauged the ability of ACT to infer an inherently sequential algorithm from statically presented data by presenting large binary vectors to the network and asking it to determine the parity.
By varying the number of binary bits for which parity must be calculated we were also able to assess ACT's ability to adapt the amount of computation to the difficulty of the vector.

The input vectors had 64 elements, of which a random number from $1$ to $64$ were randomly set to $1$ or $-1$ and the rest were set to 0.
The corresponding target was 1 if there was an odd number of ones and 0 if there was an even number of ones.
Each training sequence consisted of a single input and target vector, an example of which is shown in \fref{parity_example}.
The network architecture was a simple RNN with a single hidden layer containing 128 $tanh$ units and a single sigmoidal output unit, trained with binary cross-entropy loss on minibatches of size 128.
Note that without ACT the recurrent connection in the hidden layer was never used since the data had no sequential component, and the network reduced to a feedforward network with a single hidden layer.

\figt{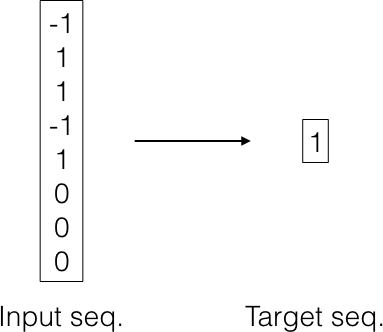}{parity_example}{0.25}{Parity training Example}{ Each sequence consists of a single input and target vector. Only 8 of the 64 input bits are shown for clarity.}

\fref{parity_bar} demonstrates that the network was unable to reliably solve the problem without ACT, with a mean of almost 40\% error compared to 50\% for random guessing.
For penalties of 0.03 and below the mean error was below 5\%.
\fref{parity_plot} reveals that the solutions were both more rapid and more accurate with lower time penalties.
It also highlights the relationship between the time penalty, the classification error rate and the average ponder time per input.
The variance in ponder time for low $\tau$ networks is very high, indicating that many correct solutions with widely varying runtime can be discovered.
We speculate that progressively higher $\tau$ values lead the network to compute the parities of successively larger chunks of the input vector at each ponder step, then iteratively combine these calculations to obtain the parity of the complete vector.

\figt{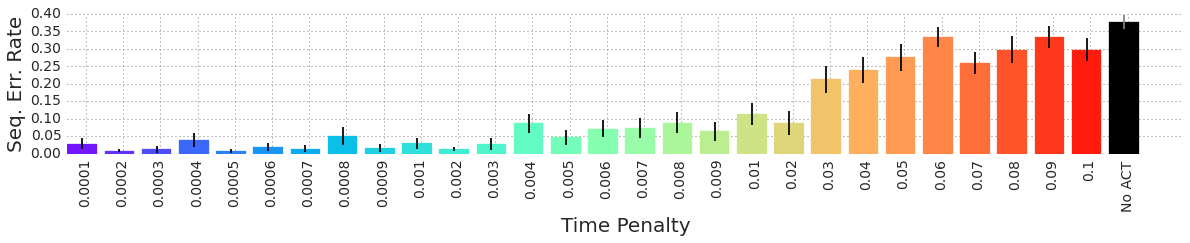}{parity_bar}{\barwidth}{Parity Error Rates}{ Bar heights show the mean error rates for different time penalties at the end of training. The error bars show the standard error in the mean.}
\figt{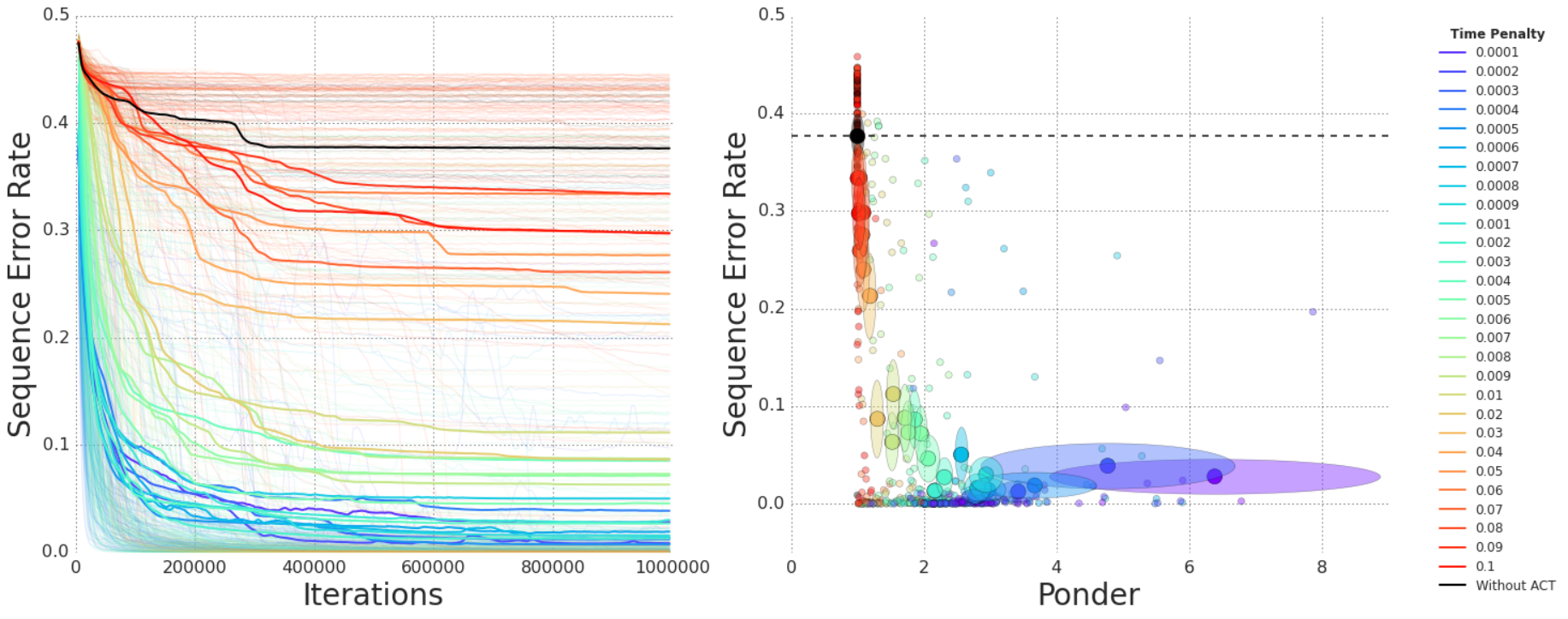}{parity_plot}{\plotwidth}{Parity Learning Curves and Error Rates Versus Ponder Time}{ \textbf{Left:} faint coloured curves show the errors for individual runs. Bold lines show the mean errors over all 20 runs for each $\tau$ value. `Iterations' is the number of gradient updates per asynchronous worker. \textbf{Right:} Small circles represent individual runs after training is complete, large circles represent the mean over 20 runs for each $\tau$ value. `Ponder' is the mean number of computation steps per input timestep (minimum 1). The black dotted line shows the mean error for the networks without ACT. The height of the ellipses surrounding the mean values represents the standard error over error rates for that value of $\tau$, while the width shows the standard error over ponder times.}

\fref{parity_difficulty} shows that for the networks without ACT and those with overly high time penalties, the error rate increases sharply with the difficulty of the task (where \emph{difficulty} is defined as the number of bits whose parity must be determined), while the amount of ponder remains roughly constant.
For the more successful networks, with intermediate $\tau$ values, ponder time appears to grow linearly with difficulty, with a slope that generally increases as $\tau$ decreases.
Even for the best networks the error rate increased somewhat with difficulty.
For some of the lowest $\tau$ networks there is a dramatic increase in ponder after about 32 bits, suggesting an inefficient algorithm.

\figt{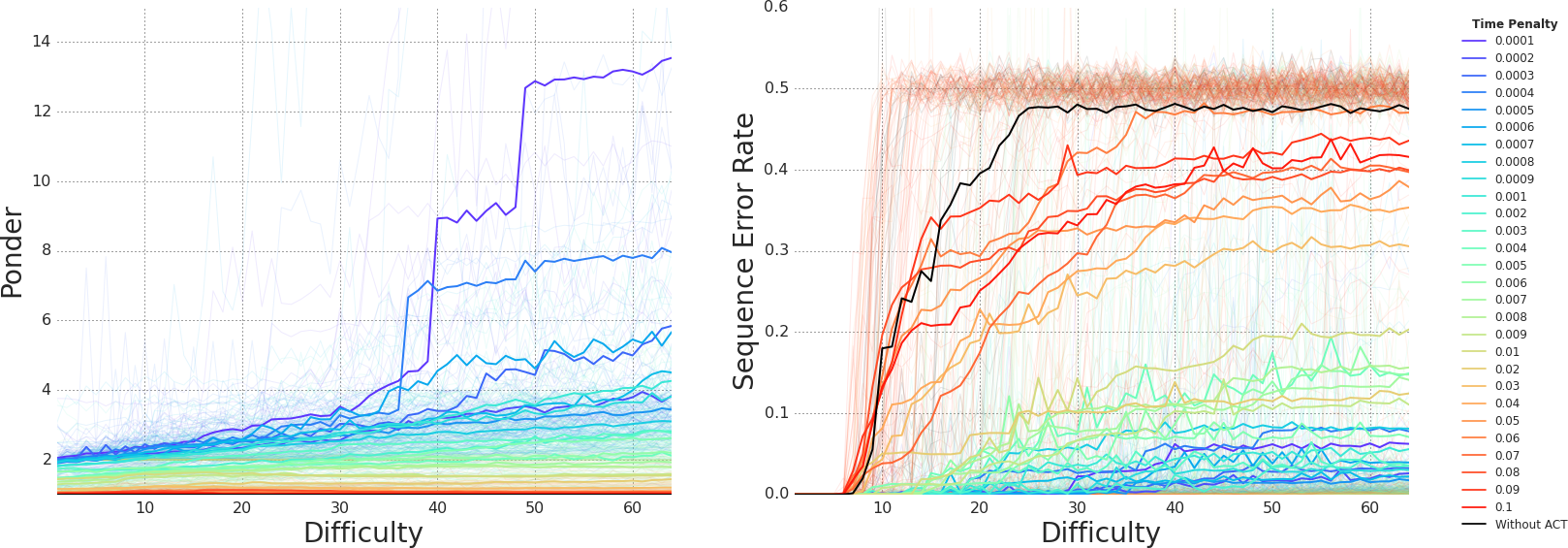}{parity_difficulty}{\plotwidth}{Parity Ponder Time and Error Rate Versus Input Difficulty}{ Faint lines are individual runs, bold lines are means over 20 networks. `Difficulty' is the number of bits in the parity vectors, with a mean over 1,000 random vectors used for each data-point.}

\subsection{Logic}
Like parity, the \emph{logic} task tests if an RNN with ACT can sequentially process a static vector.
Unlike parity it also requires the network to internally transfer information across successive input timesteps, thereby testing whether ACT can propagate coherent internal states.

Each input sequence consists of a random number from 1 to 10 of size 102 input vectors.
The first two elements of each input represent a pair of binary numbers; the remainder of the vector is divided up into 10 chunks of size 10.
The first $B$ chunks, where $B$ is a random number from 1 to 10, contain one-hot representations of randomly chosen numbers between 1 and 10; each of these numbers correspond to an index into the subset of binary logic gates whose truth tables are listed in \tref{truth}.
The remaining $10-B$ chunks were zeroed to indicate that no further binary operations were defined for that vector.
The binary target $b_{B+1}$ for each input is the truth value yielded by recursively applying the $B$ binary gates in the vector to the two initial bits $b_{1},b_{0}$. That is for $1\leq b\leq B$:
\begin{equation}\elabel{logic}
b_{i+1} = T_i(b_{i},b_{i-1}) 
\end{equation}
where $T_i(.,.)$ is the truth table indexed by chunk $i$ in the input vector.

\begin{table}
\centering
\begin{footnotesize}
\centering
\caption{Binary Truth Tables for the Logic Task}
\medskip
\tlabel{truth}
\begin{tabular}{|cc|c|c|c|c|c|c|c|c|c|c|}
\hline
\textbf{P} & \textbf{Q} & \textbf{NOR} & \textbf{Xq} & \textbf{ABJ} & \textbf{XOR} & \textbf{NAND} & \textbf{AND} & \textbf{XNOR} & \textbf{if/then} & \textbf{then/if} & \textbf{OR}\\
\hline
\textbf{T} & \textbf{T} & F & F & F & F & F & T & T & T & T & T \\
\textbf{T} & \textbf{F} & F & F & T & T & T & F & F & F & T & T \\
\textbf{F} & \textbf{T} & F & T & F & T & T & F & F & T & F & T \\
\textbf{F} & \textbf{F} & T & F & F & F & T & F & T & T & T & F \\
\hline
\end{tabular}
\end{footnotesize}
\end{table}

For the first vector in the sequence, the two input bits $b_0,b_1$ were randomly chosen to be false (0) or true (1) and assigned to the first two elements in the vector.
For subsequent vectors, only $b_1$ was random, while $b_0$ was implicitly equal to the target bit from the previous vector (for the purposes of calculating the current target bit), but was always set to zero in the input vector.
To solve the task, the network therefore had to learn both how to calculate the sequence of binary operations represented by the chunks in each vector, and how to carry the final output of that sequence over to the next timestep. 
An example input-target sequence pair is shown in \fref{logic_example}.

\figt{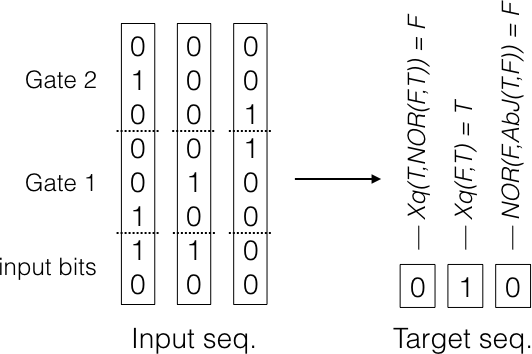}{logic_example}{0.4}{Logic training Example}{ Both the input and target sequences consist of 3 vectors. For simplicity only 2 of the 10 possible logic gates represented in the input are shown, and each is restricted to one of the first 3 gates in \tref{truth} (NOR, Xq, and ABJ). The segmentation of the input vectors is show on the left and the recursive application of \eref{logic} required to determine the targets (and subsequent $b_0$ values) is shown in italics above the target vectors.}

The network architecture was single-layer LSTM with 128 cells.
The output was a single sigmoidal unit, trained with binary cross-entropy, and the minibatch size was 16.

\fref{logic_bar} shows that the network reaches a minimum sequence error rate of around 0.2 without ACT (compared to 0.5 for random guessing), and virtually zero error for all $\tau \leq 0.01$.
From \fref{logic_plot} it can be seen that low $\tau$ ACT networks solve the task very quickly, requiring about 10,000 training iterations.
For higher $\tau$ values ponder time reduces to 1, at which point the networks trained with ACT behave identically to those without.
For lower $\tau$ values, the spread of ponder values, and hence computational cost, is quite large. 
Again we speculate that this is due to the network learning more or less `chunked' solutions in which composite truth table are learned for multiple successive logic operations.
This is somewhat supported by the clustering of the lowest $\tau$ networks around a ponder time of 5--6, which is approximately the mean number of logic gates applied per sequence, and hence the minimum number of computations the network would need if calculating single binary operations at a time.

\figt{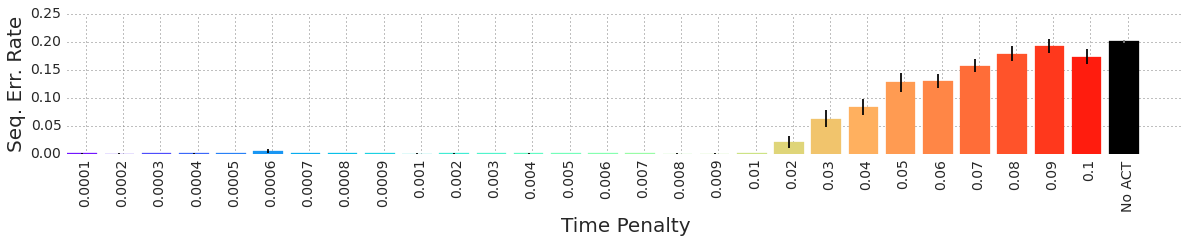}{logic_bar}{\barwidth}{Logic Error Rates}{}
\figt{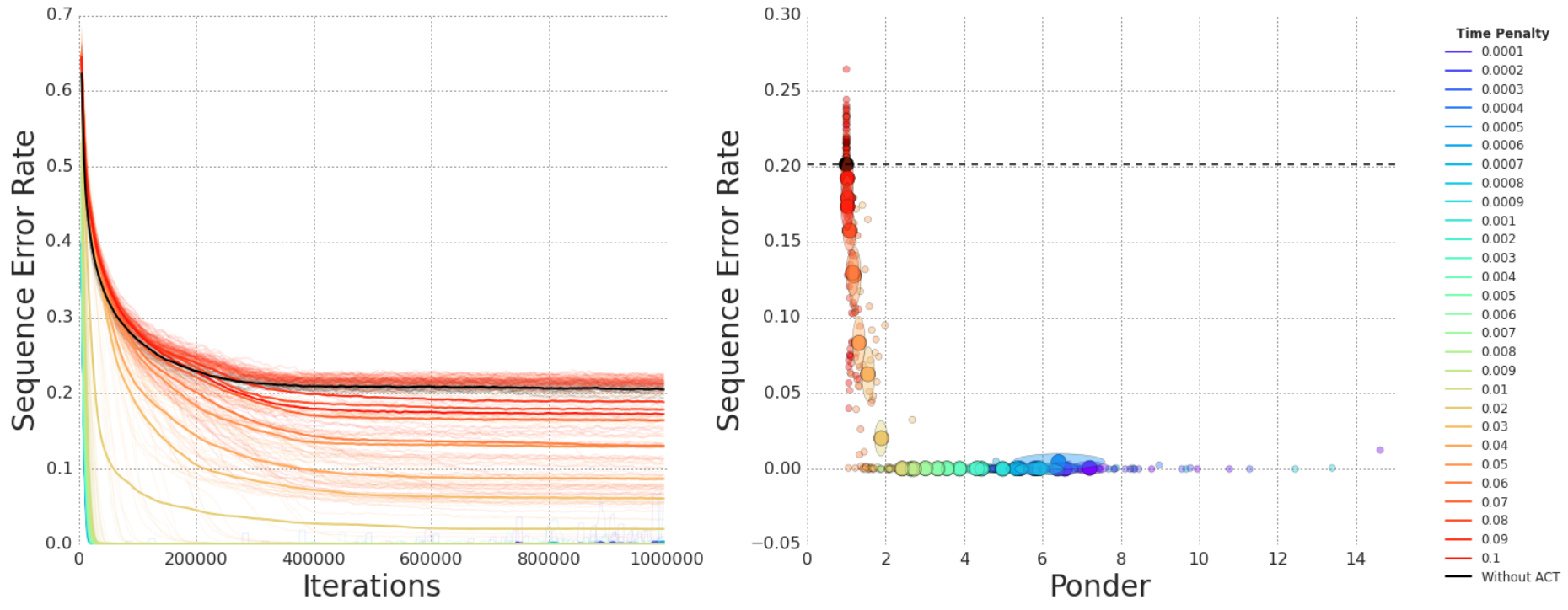}{logic_plot}{\plotwidth}{Logic Learning Curves and Error Rates Versus Ponder Time}{}

\fref{logic_difficulty} shows a surprisingly high ponder time for the least difficult inputs, with some networks taking more than 10 steps to evaluate a single logic gate.
From 5 to 10 logic gates, ponder gradually increases with difficulty as expected, suggesting that a qualitatively different solution is learned for the two regimes.
This is supported by the error rates for the non ACT and high $\tau$ networks, which increase abruptly after 5 gates.
It may be that 5 is the upper limit on the number of successive gates the network can learn as a single composite operation, and thereafter it is forced to apply an iterative algorithm.

\figt{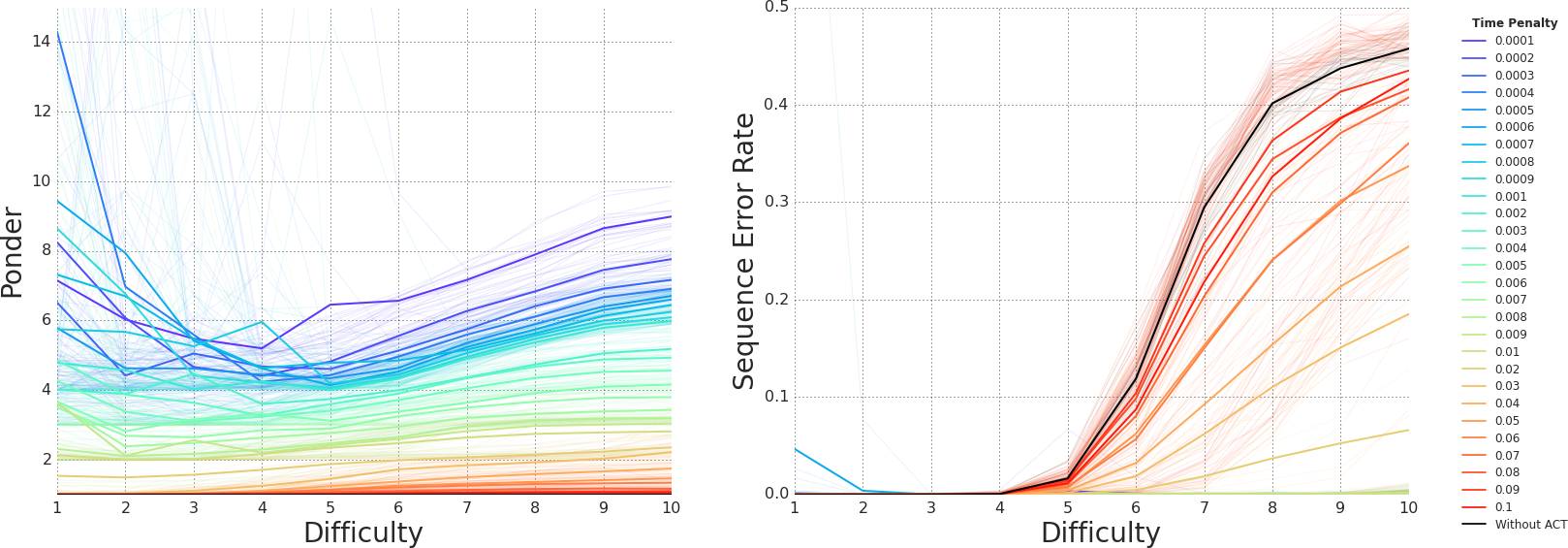}{logic_difficulty}{\plotwidth}{Logic Ponder Time and Error Rate Versus Input Difficulty}{ `Difficulty' is the number of logic gates in each input vector; all sequences were length 5.}

\subsection{Addition}
The addition task presents the network with a input sequence of 1 to 5 size 50 input vectors.
Each vector represents a $D$ digit number, where $D$ is drawn randomly from 1 to 5, and each digit is drawn randomly from 0 to 9.
The first $10D$ elements of the vector are a concatenation of one-hot encodings of the $D$ digits in the number, and the remainder of the vector is set to 0.
The required output is the cumulative sum of all inputs up to the current one, represented as a set of 6 simultaneous classifications for the 6 possible digits in the sum.
There is no target for the first vector in the sequence, as no sums have yet been calculated.
Because the previous sum must be carried over by the network, this task again requires the internal state of the network to remain coherent.
Each classification is modelled by a size 11 softmax, where the first 10 classes are the digits and the $11^{th}$ is a special marker used to indicate that the number is complete.
An example input-target pair is shown in \fref{addition_example}.

\figt{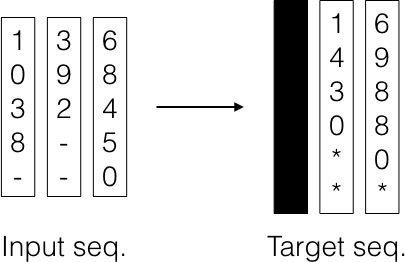}{addition_example}{0.3}{Addition training Example}{ Each digit in the input sequence is represented by a size 10 one hot encoding. Unused input digits, marked `-', are represented by a vector of 10 zeros. The black vector at the start of the target sequence indicates that no target was required for that step. The target digits are represented as 1-of-11 classes, where the $11^th$ class, marked `*', is used for digits beyond the end of the target number.}

The network was single-layer LSTM with 512 memory cells.
The loss function was the joint cross-entropy of all 6 targets at each time-step where targets were present and the minibatch size was 32.
The maximum ponder $M$ was set to 20 for this task, as it was found that some networks had very high ponder times early in training.

The results in \fref{addition_bar} show that the task was perfectly solved by the ACT networks for all values of $\tau$ in the grid search.
Unusually, networks with higher $\tau$ solved the problem with fewer training examples. 
\fref{addition_difficulty} demonstrates that the relationship between the ponder time and the number of digits was approximately linear for most of the ACT networks, and that for the most efficient networks (with the highest $\tau$ values) the slope of the line was close to 1, which matches our expectations that an efficient long addition algorithm should need one computation step per digit.

\figt{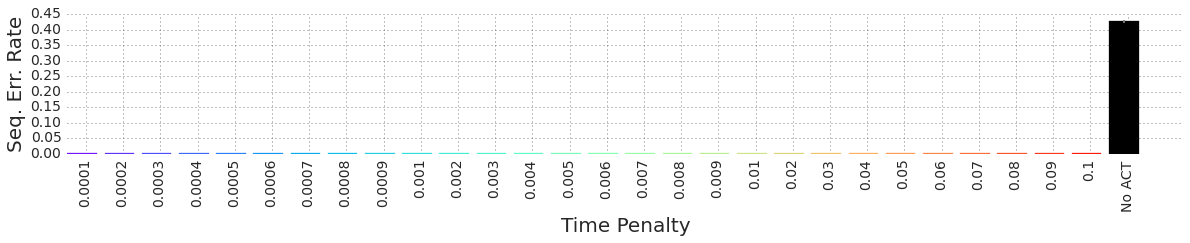}{addition_bar}{\barwidth}{Addition Error Rates}{}
\figt{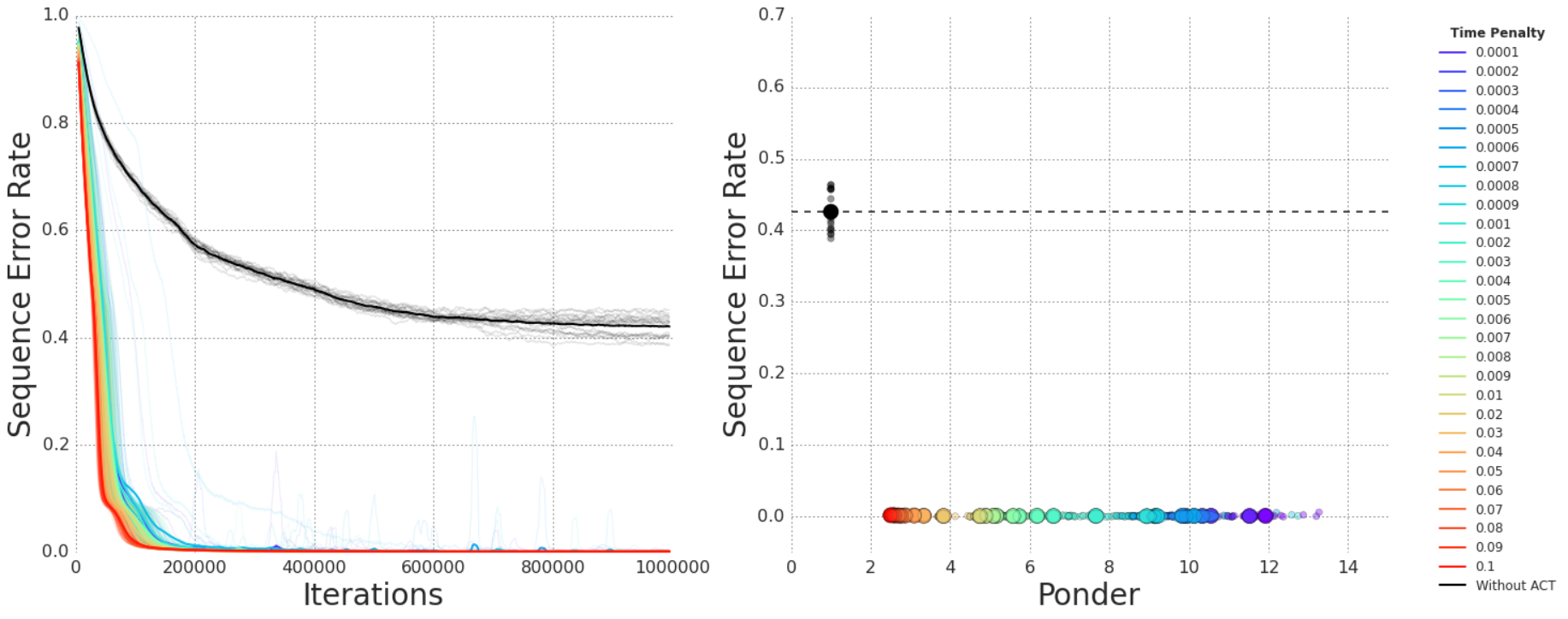}{addition_plot}{\plotwidth}{Addition Learning Curves and Error Rates Versus Ponder Time}{}
\figt{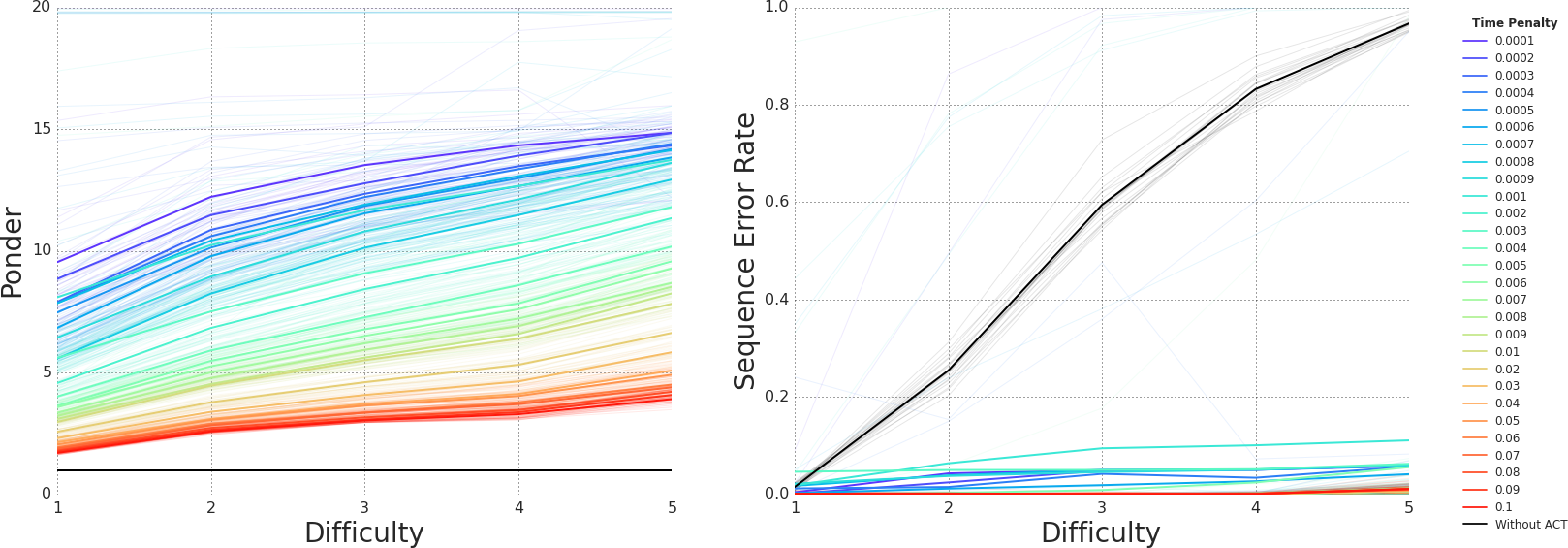}{addition_difficulty}{\plotwidth}{Addition Ponder Time and Error Rate Versus Input Difficulty}{ `Difficulty' is the number of digits in each input vector; all sequences were length 3.}

\fref{addition_sequence} shows how the ponder time is distributed during individual addition sequences, providing further evidence of an approximately linear-time long addition algorithm.  
\figt{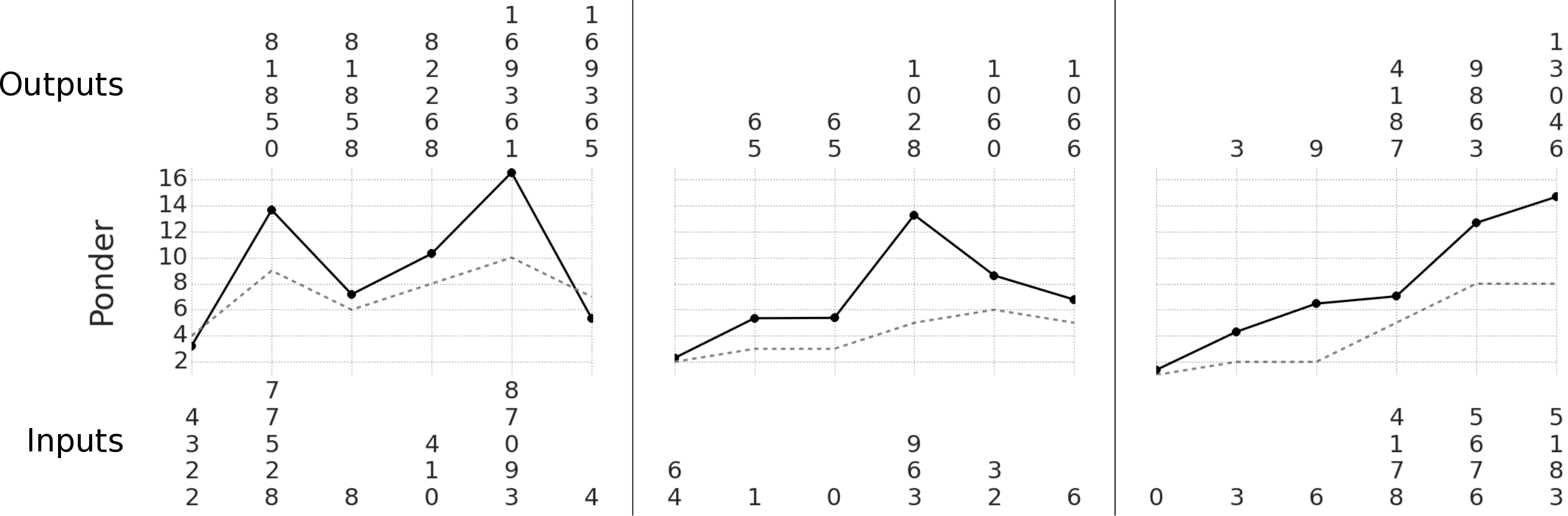}{addition_sequence}{\plotwidth}{Ponder Time During Three Addition Sequences}{ The input sequence is shown along the bottom x-axis and the network output sequence is shown along the top x-axis. The ponder time $\rho_t$ at each input step is shown by the black lines; the actual number of computational steps taken at each point is $\rho_t$ rounded up to the next integer. The grey lines show the total number of digits in the two numbers being summed at each step; this appears to give a rough lower bound on the ponder time, suggesting an internal algorithm that is approximately linear in the number of digits. All plots were created using the same network, trained with $\tau=9e^{-4}$.}

\subsection{Sort}
The \emph{sort} task requires the network to sort sequences of 2 to 15 numbers drawn from a standard normal distribution in ascending order.
The experiments considered so far have been designed to favour ACT by compressing sequential information into single vectors, and thereby requiring the use of multiple computation steps to unpack them.
For the sort task a more natural sequential representation was used: the random numbers were presented one at a time as inputs, and the required output was the sequence of indices into the number sequence placed in sorted order; an example is shown in \fref{sort_example}.
We were particularly curious to see how the number of ponder steps scaled with the number of elements to be sorted, knowing that efficient sorting algorithms have $O(N\log N)$ computational cost.

\figt{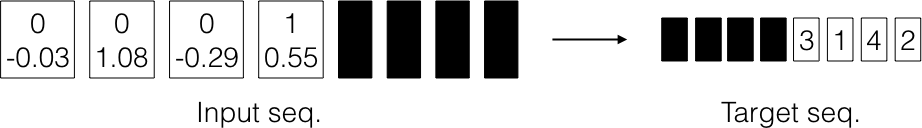}{sort_example}{0.6}{Sort training Example}{ Each size 2 input vector consists of one real number and one binary flag to indicate the end of sequence to be sorted; inputs following the sort sequence are set to zero and marked in black. No targets are present until after the sort sequence; thereafter the size 15 target vectors represent the sorted indices of the input sequence.}

The network was single-layer LSTM with 512 cells.
The output layer was a size 15 softmax, trained with cross-entropy to classify the indices of the sorted inputs.
The minibatch size was 16.

\fref{sort_bar} shows that the advantage of using ACT is less dramatic for this task than the previous three, but still substantial (from around 12\% error without ACT to around 6\% for the best $\tau$ value).
However from \fref{sort_plot} it is clear that these gains come at a heavy computational cost, with the best networks requiring roughly 9 times as much computation as those without ACT.
Not surprisingly, \fref{sort_difficulty} shows that the error rate grew rapidly with the sequence length for all networks.
It also indicates that the better networks had a sublinear growth in computations per input step with sequence length, though whether this indicates a logarithmic time algorithm is unclear.
One problem with the sort task was that the Gaussian samples were sometimes very close together, making it hard for the network to determine which was greater; enforcing a minimum separation between successive values would probably be beneficial.

\figt{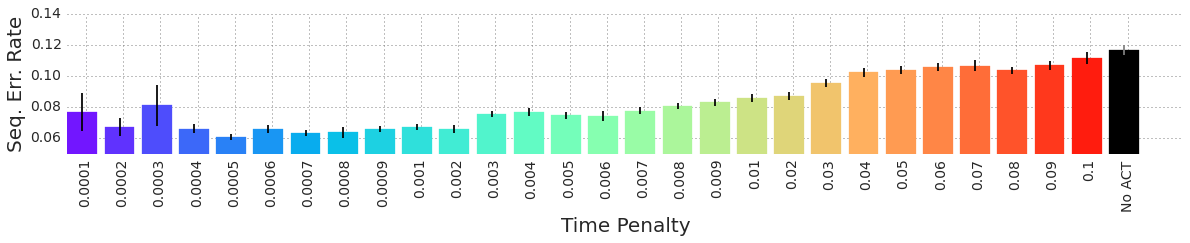}{sort_bar}{\barwidth}{Sort Error Rates}{}
\figt{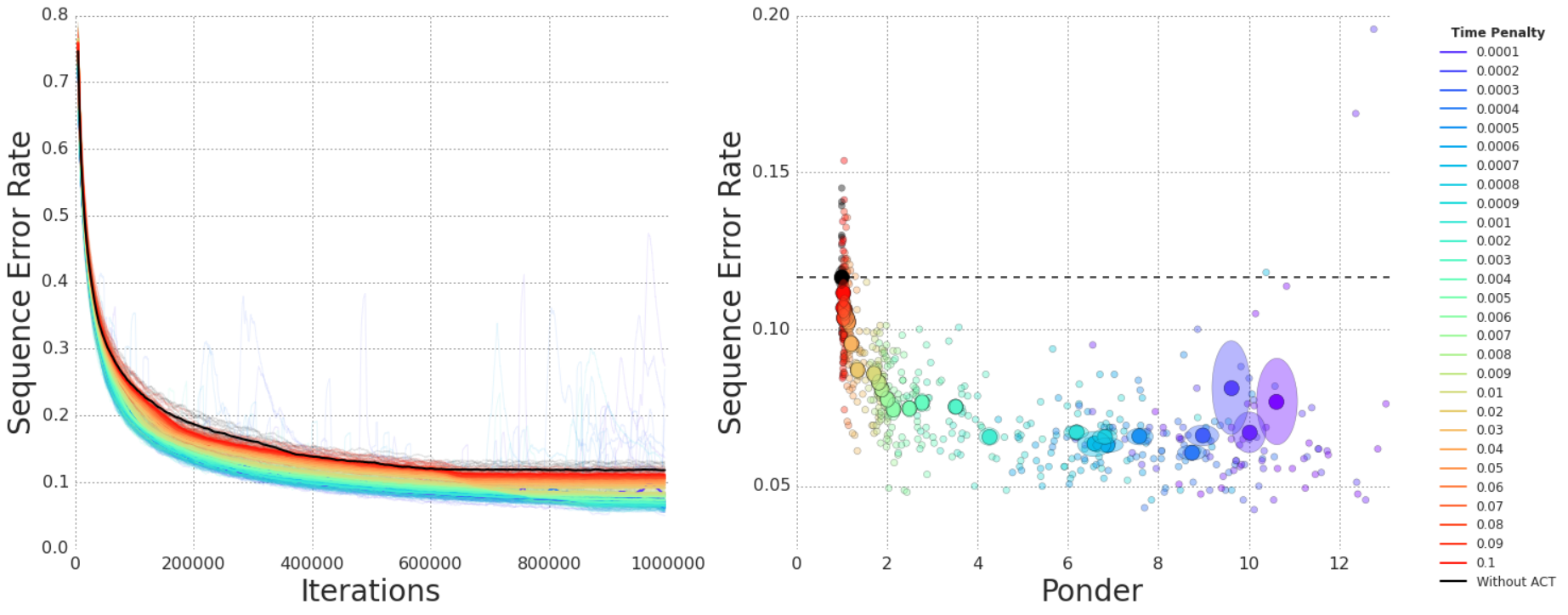}{sort_plot}{\plotwidth}{Sort Learning Curves and Error Rates Versus Ponder Time}{}
\figt{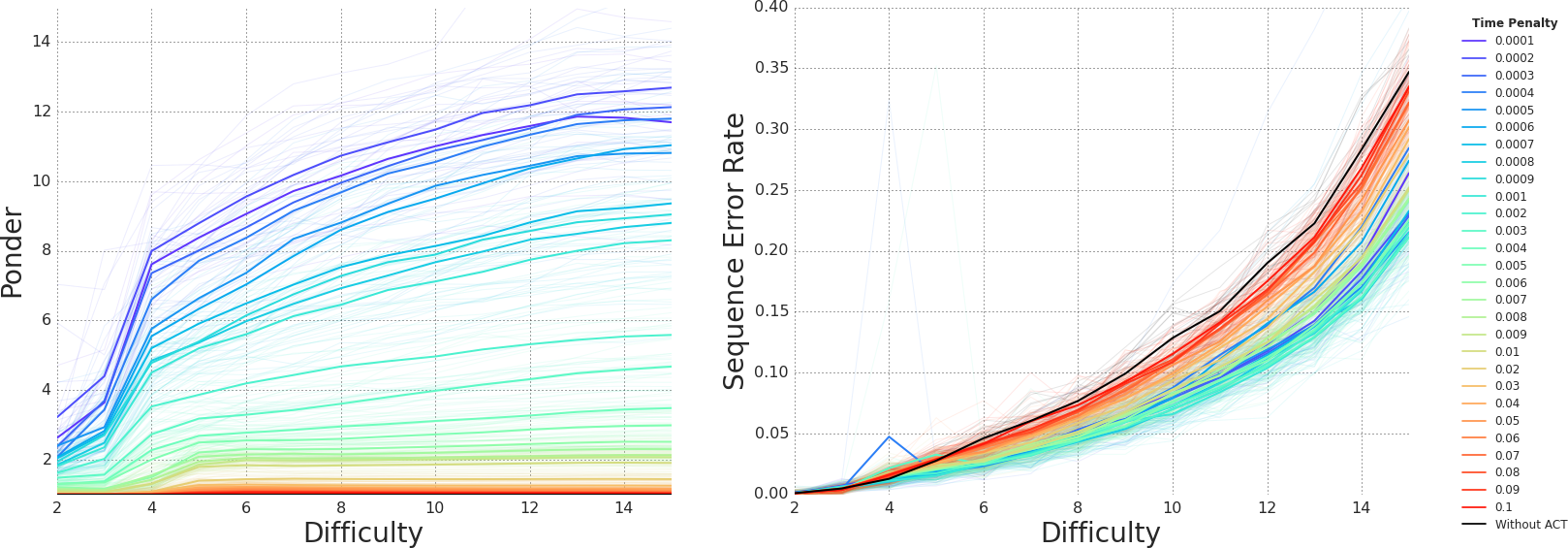}{sort_difficulty}{\plotwidth}{Sort Ponder Time and Error Rate Versus Input Difficulty}{ `Difficulty' is the length of the sequence to be sorted.}

\fref{sort_sequence} shows the ponder time during three sort sequences of varying length. As can be seen, there is a large spike in ponder time near (though not precisely at) the end of the input sequence, presumably when the majority of the sort comparisons take place. Note that the spike is much higher for the longer two sequences than the length 5 one, again pointing to an algorithm that is nonlinear in sequence length (the average ponder per timestep is nonetheless lower for longer sequences, as little pondering is done away from the spike.). 

\figt{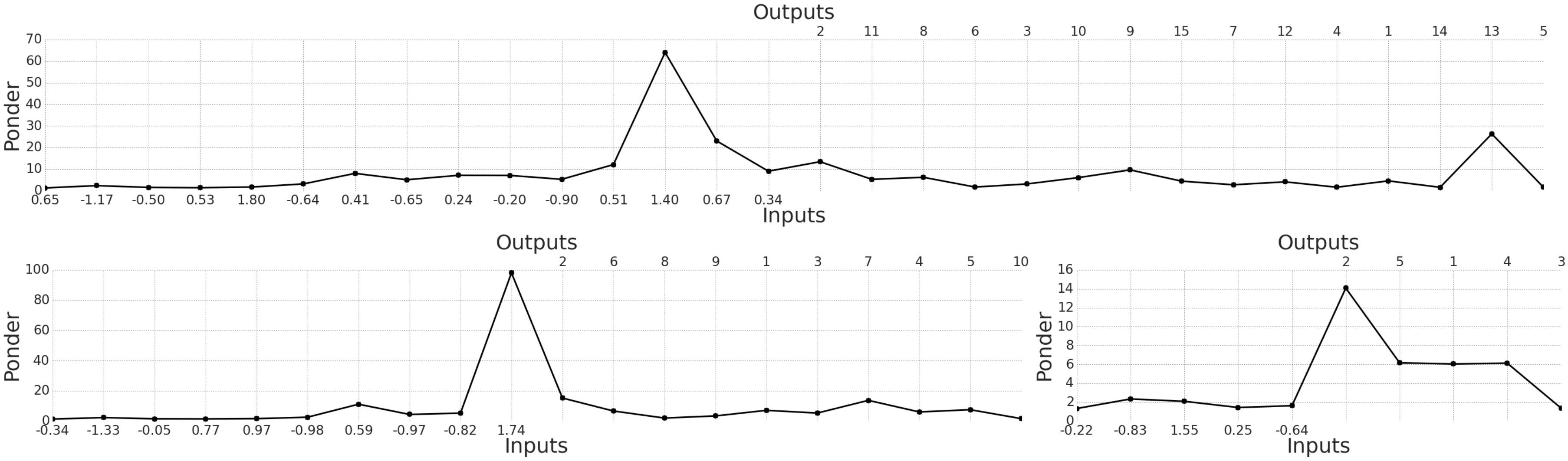}{sort_sequence}{\plotwidth}{Ponder Time During Three Sort Sequences}{ The input sequences to be sorted are shown along the bottom x-axes and the network output sequences are shown along the top x-axes. 
All plots created using the same network, trained with $\tau=10^{-3}$.}

\subsection{Wikipedia Character Prediction}
The \emph{Wikipedia} task is character prediction on text drawn from the Hutter prize Wikipedia dataset~\cite{hutter2005universal}.
Following previous RNN experiments on the same data~\cite{graves2013generating}, the raw unicode text was used, including XML tags and markup characters, with one byte presented per input timestep and the next byte predicted as a target.
No validation set was used for early stopping, as the networks were unable to overfit the data, and all error rates are recorded on the training set.
Sequences of 500 consecutive bytes were randomly chosen from the training set and presented to the network, whose internal state was reset to 0 at the start of each sequence.

LSTM networks were used with a single layer of 1500 cells and a size 256 softmax classification layer.
As can be seen from \fref{enwik_bar,fig:enwik_plot}, the error rates are fairly similar with and without ACT, and across values of $\tau$ (although the learning curves suggest that the ACT networks are somewhat more data efficient).
Furthermore the amount of ponder per input is much lower than for the other problems, suggesting that the advantages of extra computation were slight for this task.

\figt{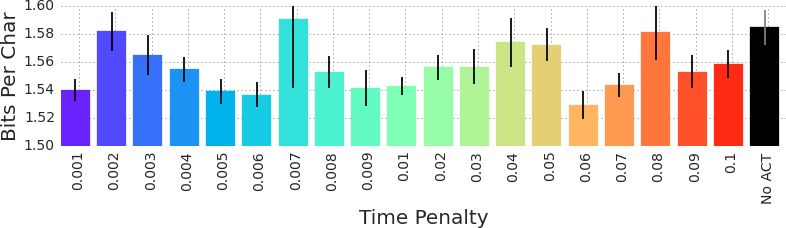}{enwik_bar}{0.7}{Wikipedia Error Rates}{}
\figt{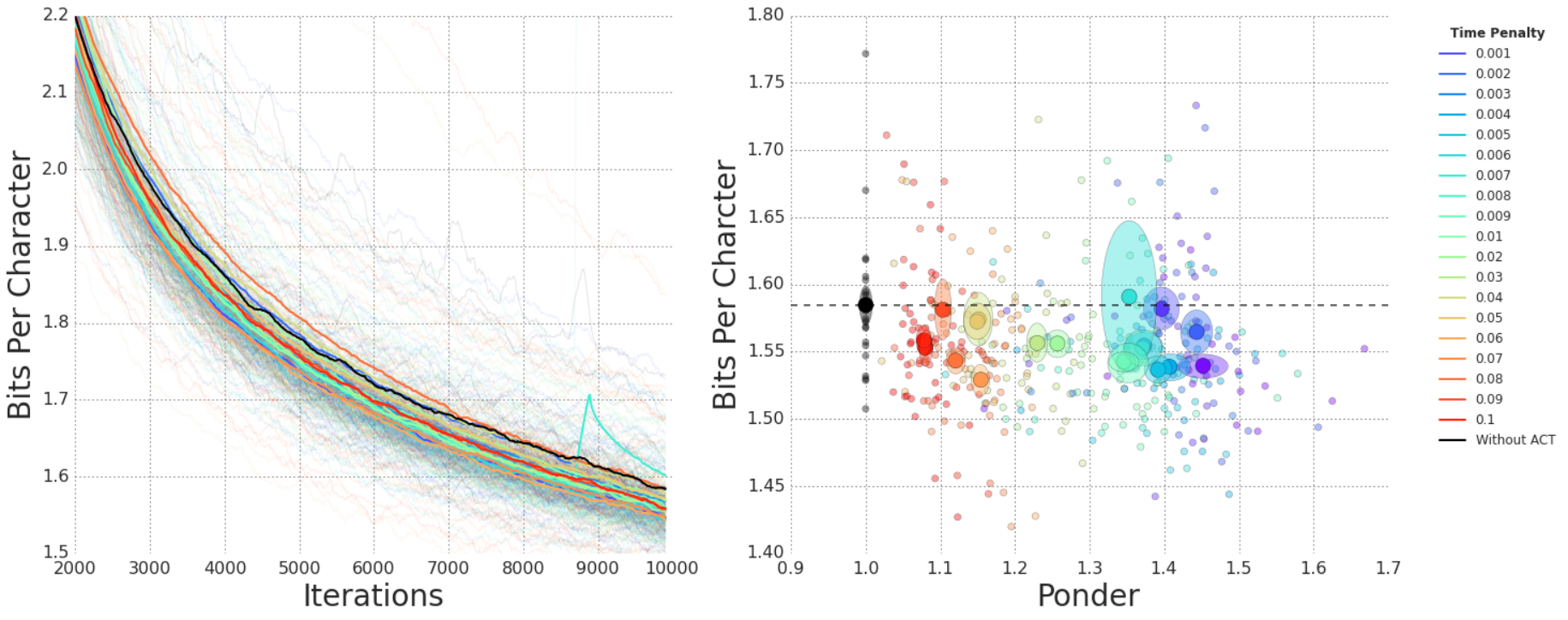}{enwik_plot}{\plotwidth}{Wikipedia Learning Curves (Zoomed) and Error Rates Versus Ponder Time}{}

However \fref{enwik_sequence} reveals an intriguing pattern of ponder allocation while processing a sequence. 
Character prediction networks trained with ACT consistently pause at spaces between words, and pause for longer at `boundary' characters such as commas and full stops.
We speculate that the extra computation is used to make predictions about the next `chunk' in the data (word, sentence, clause), much as humans have been found to do in self-paced reading experiments~\cite{just1982paradigms}.
This suggests that ACT could be useful for inferring implicit boundaries or transitions in sequence data.
Alternative measures for inferring transitions include the next-step prediction loss and predictive entropy, both of which tend to increase during harder predictions.
However, as can be seen from the figure, they are a less reliable indicator of boundaries, and are not likely to increase at points such as full stops and commas, as these are invariably followed by space characters.
More generally, loss and entropy only indicate the difficulty of the current prediction, not the degree to which the current input is likely to impact future predictions.

\figt{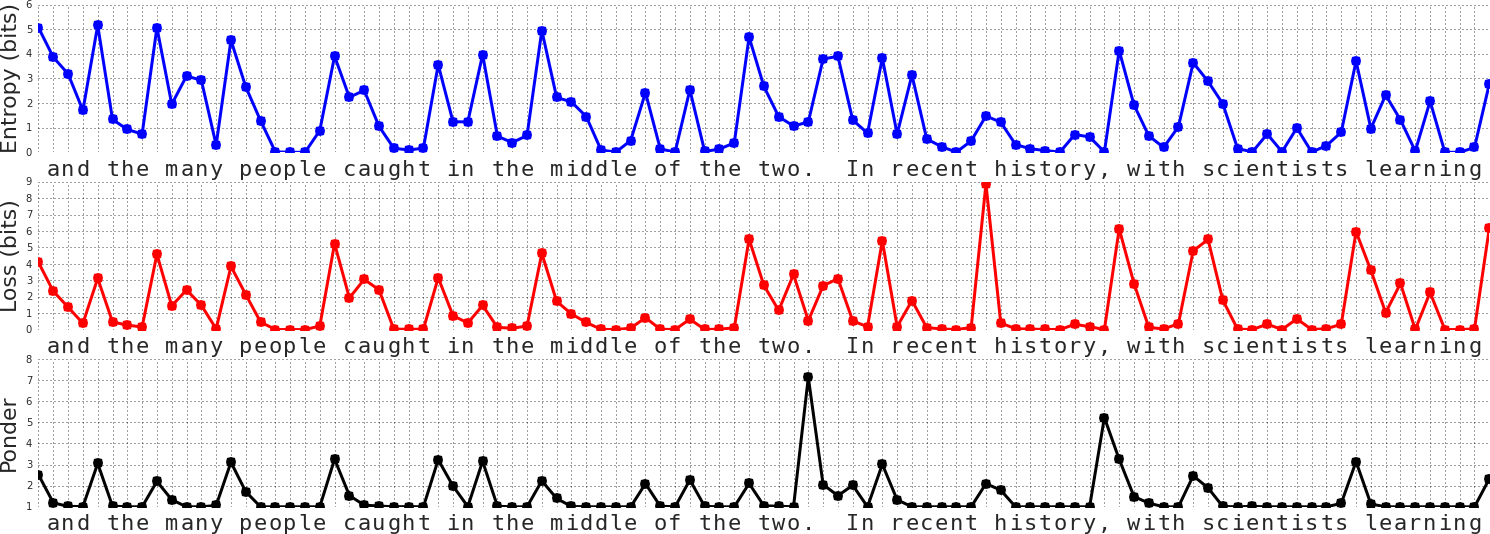}{enwik_sequence}{\plotwidth}{Ponder Time, Prediction loss and Prediction Entropy During a Wikipedia Text Sequence}{ Plot created using a network trained with $\tau = 6e^{-3}$}

Furthermore \fref{enwik_sequence_2} reveals that, as well as being an effective detector of non-text transition markers such as the opening brackets of XML tags, ACT does not increase computation time during random or fundamentally unpredictable sequences like the two ID numbers.
This is unsurprising, as doing so will not improve its predictions.
In contrast, both entropy and loss are inevitably high for unpredictable data.
We are therefore hopeful that computation time will provide a better way to distinguish between structure and noise (or at least data perceived by the network as structure or noise) than existing measures of predictive difficulty.

\figt{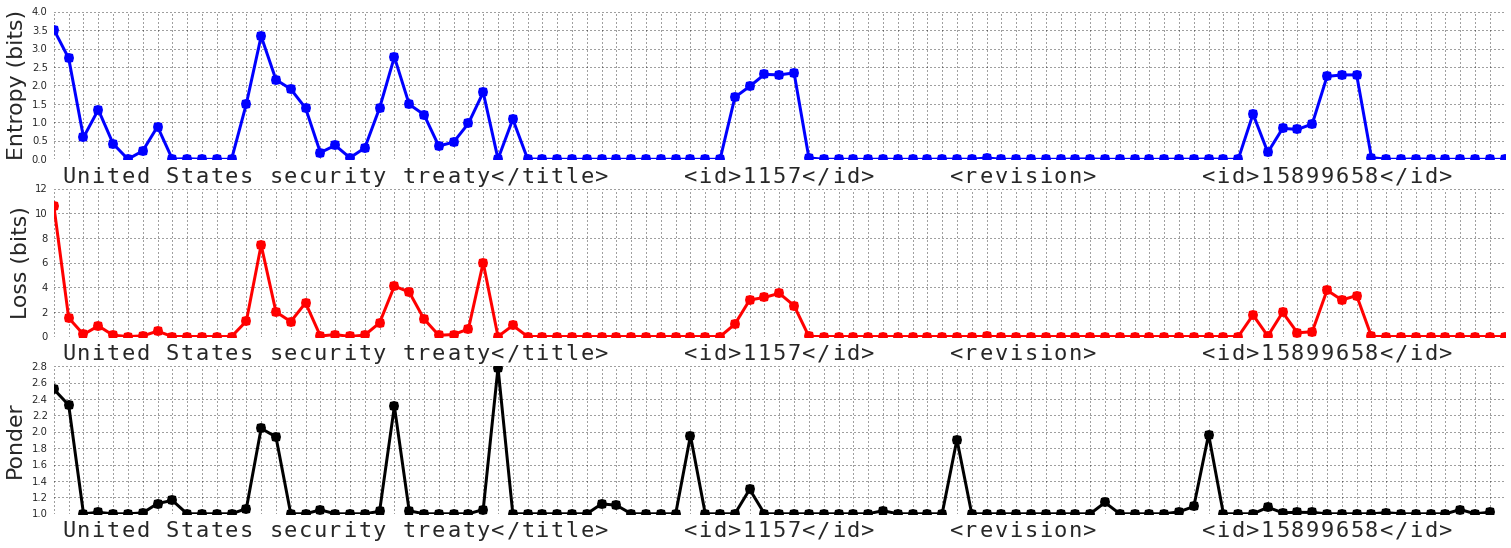}{enwik_sequence_2}{\plotwidth}{Ponder Time, Prediction loss and Prediction Entropy During a Wikipedia Sequence Containing XML Tags}{ Created using the same network as \fref{enwik_sequence}.}

\section{Conclusion}\seclabel{conclusion}
This paper has introduced Adaptive Computation time (ACT), a method that allows recurrent neural networks to learn how many updates to perform for each input they receive.
Experiments on synthetic data prove that ACT can make otherwise inaccessible problems straightforward for RNNs to learn, and that it is able to dynamically adapt the amount of computation it uses to the demands of the data.
An experiment on real data suggests that the allocation of computation steps learned by ACT can yield insight into both the structure of the data and the computational demands of predicting it.

ACT promises to be particularly interesting for recurrent architectures containing soft attention modules~\cite{BahdanauCB14,graves2014ntm,vinyals2015pointer,gregor2015draw}, which it could enable to dynamically adapt the number of glances or internal operations they perform at each time-step.

One weakness of the current algorithm is that it is quite sensitive to the time penalty parameter that controls the relative cost of computation time versus prediction error.
An important direction for future work will be to find ways of automatically determining and adapting the trade-off between accuracy and speed.

\section*{Acknowledgments} 
The author wishes to thank Ivo Danihleka, Greg Wayne, Tim Harley, Malcolm Reynolds, Jacob Menick, Oriol Vinyals, Joel Leibo, Koray Kavukcuoglu and many others on the DeepMind team for valuable comments and suggestions, as well as Albert Zeyer, Martin Abadi, Dario Amodei, Eugene Brevdo and Christopher Olah for pointing out the discontinuity in the ponder cost, which was erroneously described as smooth in an earlier version of the paper.

\bibliographystyle{abbrv}
\bibliography{act_arxiv} 

\begin{thebibliography}{10}

\bibitem{an1996effects}
G.~An.
\newblock The effects of adding noise during backpropagation training on a
  generalization performance.
\newblock {\em Neural Computation}, 8(3):643--674, 1996.

\bibitem{BahdanauCB14}
D.~Bahdanau, K.~Cho, and Y.~Bengio.
\newblock Neural machine translation by jointly learning to align and
  translate.
\newblock abs/1409.0473, 2014.

\bibitem{bengio2015conditional}
E.~Bengio, P.-L. Bacon, J.~Pineau, and D.~Precup.
\newblock Conditional computation in neural networks for faster models.
\newblock {\em arXiv preprint arXiv:1511.06297}, 2015.

\bibitem{ciresan2012multicolumn}
D.~C. Ciresan, U.~Meier, and J.~Schmidhuber.
\newblock Multi-column deep neural networks for image classification.
\newblock In {\em arXiv:1202.2745v1 [cs.CV]}, 2012.

\bibitem{dahl2012speech}
G.~Dahl, D.~Yu, L.~Deng, and A.~Acero.
\newblock Context-dependent pre-trained deep neural networks for
  large-vocabulary speech recognition.
\newblock {\em Audio, Speech, and Language Processing, IEEE Transactions on},
  20(1):30 --42, jan. 2012.

\bibitem{denoyer2014deep}
L.~Denoyer and P.~Gallinari.
\newblock Deep sequential neural network.
\newblock {\em arXiv preprint arXiv:1410.0510}, 2014.

\bibitem{eslami2016attend}
S.~Eslami, N.~Heess, T.~Weber, Y.~Tassa, K.~Kavukcuoglu, and G.~E. Hinton.
\newblock Attend, infer, repeat: Fast scene understanding with generative
  models.
\newblock {\em arXiv preprint arXiv:1603.08575}, 2016.

\bibitem{graves2013generating}
A.~Graves.
\newblock Generating sequences with recurrent neural networks.
\newblock {\em arXiv preprint arXiv:1308.0850}, 2013.

\bibitem{graves2013speech}
A.~Graves, A.~Mohamed, and G.~Hinton.
\newblock Speech recognition with deep recurrent neural networks.
\newblock In {\em Acoustics, Speech and Signal Processing (ICASSP), 2013 IEEE
  International Conference on}, pages 6645--6649. IEEE, 2013.

\bibitem{graves2014ntm}
A.~Graves, G.~Wayne, and I.~Danihelka.
\newblock Neural turing machines.
\newblock {\em arXiv preprint arXiv:1410.5401}, 2014.

\bibitem{grefenstette2015learning}
E.~Grefenstette, K.~M. Hermann, M.~Suleyman, and P.~Blunsom.
\newblock Learning to transduce with unbounded memory.
\newblock In {\em Advances in Neural Information Processing Systems}, pages
  1819--1827, 2015.

\bibitem{gregor2015draw}
K.~Gregor, I.~Danihelka, A.~Graves, and D.~Wierstra.
\newblock Draw: A recurrent neural network for image generation.
\newblock {\em arXiv preprint arXiv:1502.04623}, 2015.

\bibitem{hochreiter2001gradient}
S.~Hochreiter, Y.~Bengio, P.~Frasconi, and J.~Schmidhuber.
\newblock Gradient flow in recurrent nets: the difficulty of learning long-term
  dependencies, 2001.

\bibitem{hochreiter1997lstm}
S.~Hochreiter and J.~Schmidhuber.
\newblock Long short-term memory.
\newblock {\em Neural computation}, 9(8):1735--1780, 1997.

\bibitem{hutter2005universal}
M.~Hutter.
\newblock {\em Universal artificial intelligence}.
\newblock Springer, 2005.

\bibitem{just1982paradigms}
M.~A. Just, P.~A. Carpenter, and J.~D. Woolley.
\newblock Paradigms and processes in reading comprehension.
\newblock {\em Journal of experimental psychology: General}, 111(2):228, 1982.

\bibitem{kalchbrenner2015grid}
N.~Kalchbrenner, I.~Danihelka, and A.~Graves.
\newblock Grid long short-term memory.
\newblock {\em arXiv preprint arXiv:1507.01526}, 2015.

\bibitem{kingma2014adam}
D.~Kingma and J.~Ba.
\newblock Adam: A method for stochastic optimization.
\newblock {\em arXiv preprint arXiv:1412.6980}, 2014.

\bibitem{krizhevsky2012imagenet}
A.~Krizhevsky, I.~Sutskever, and G.~E. Hinton.
\newblock Imagenet classification with deep convolutional neural networks.
\newblock In {\em Advances in neural information processing systems}, pages
  1097--1105, 2012.

\bibitem{le2014distributed}
Q.~V. Le and T.~Mikolov.
\newblock Distributed representations of sentences and documents.
\newblock {\em arXiv preprint arXiv:1405.4053}, 2014.

\bibitem{li2013introduction}
M.~Li and P.~Vit{\'a}nyi.
\newblock {\em An introduction to Kolmogorov complexity and its applications}.
\newblock Springer Science \& Business Media, 2013.

\bibitem{mikolov2013distributed}
T.~Mikolov, I.~Sutskever, K.~Chen, G.~S. Corrado, and J.~Dean.
\newblock Distributed representations of words and phrases and their
  compositionality.
\newblock In {\em Advances in neural information processing systems}, pages
  3111--3119, 2013.

\bibitem{olshausen1996emergence}
B.~A. Olshausen et~al.
\newblock Emergence of simple-cell receptive field properties by learning a
  sparse code for natural images.
\newblock {\em Nature}, 381(6583):607--609, 1996.

\bibitem{recht2011hogwild}
B.~Recht, C.~Re, S.~Wright, and F.~Niu.
\newblock Hogwild: A lock-free approach to parallelizing stochastic gradient
  descent.
\newblock In {\em Advances in Neural Information Processing Systems}, pages
  693--701, 2011.

\bibitem{reed2015npi}
S.~Reed and N.~de~Freitas.
\newblock Neural programmer-interpreters.
\newblock Technical Report arXiv:1511.06279, 2015.

\bibitem{schmidhuber2012self}
J.~Schmidhuber.
\newblock Self-delimiting neural networks.
\newblock {\em arXiv preprint arXiv:1210.0118}, 2012.

\bibitem{schmidhuber96guessing}
J.~Schmidhuber and S.~Hochreiter.
\newblock Guessing can outperform many long time lag algorithms.
\newblock Technical report, 1996.

\bibitem{srivastava2014dropout}
N.~Srivastava, G.~Hinton, A.~Krizhevsky, I.~Sutskever, and R.~Salakhutdinov.
\newblock Dropout: A simple way to prevent neural networks from overfitting.
\newblock {\em The Journal of Machine Learning Research}, 15(1):1929--1958,
  2014.

\bibitem{srivastava2015training}
R.~K. Srivastava, K.~Greff, and J.~Schmidhuber.
\newblock Training very deep networks.
\newblock In {\em Advances in Neural Information Processing Systems}, pages
  2368--2376, 2015.

\bibitem{srivastava2013first}
R.~K. Srivastava, B.~R. Steunebrink, and J.~Schmidhuber.
\newblock First experiments with powerplay.
\newblock {\em Neural Networks}, 41:130--136, 2013.

\bibitem{sukhbaatar2015end}
S.~Sukhbaatar, J.~Weston, R.~Fergus, et~al.
\newblock End-to-end memory networks.
\newblock In {\em Advances in Neural Information Processing Systems}, pages
  2431--2439, 2015.

\bibitem{sutskever2014sequence}
I.~Sutskever, O.~Vinyals, and Q.~V. Le.
\newblock Sequence to sequence learning with neural networks.
\newblock {\em arXiv preprint arXiv:1409.3215}, 2014.

\bibitem{vinyals2015order}
O.~Vinyals, S.~Bengio, and M.~Kudlur.
\newblock Order matters: Sequence to sequence for sets.
\newblock {\em arXiv preprint arXiv:1511.06391}, 2015.

\bibitem{vinyals2015pointer}
O.~Vinyals, M.~Fortunato, and N.~Jaitly.
\newblock Pointer networks.
\newblock In {\em Advances in Neural Information Processing Systems}, pages
  2674--2682, 2015.

\bibitem{wiles95fermat}
A.~J. Wiles.
\newblock Modular elliptic curves and fermat’s last theorem.
\newblock {\em ANNALS OF MATH}, 141:141, 1995.

\bibitem{williams1995gradient}
R.~J. Williams and D.~Zipser.
\newblock Gradient-based learning algorithms for recurrent networks and their
  computational complexity.
\newblock {\em Back-propagation: Theory, architectures and applications}, pages
  433--486, 1995.

\end{thebibliography}
                                                        
\end{document}